\documentclass[sigconf]{acmart}

\usepackage{cleveref}
\usepackage{graphicx}
\usepackage{lscape}
\usepackage{blindtext}
\usepackage{hyperref}

\newtheorem{theorem}{Theorem}[section]
\newtheorem{lemma}[theorem]{Lemma}

\theoremstyle{definition}

\usepackage{multirow}

\newcommand{\norm}[1]{\left\lVert#1\right\rVert}
\newcommand{\tr}{\operatorname{Tr}}
\newcommand{\inner}[2]{\langle #1, #2 \rangle}
\newcommand{\fidelity}{\textit{fidelity }}

\AtBeginDocument{%
  \providecommand\BibTeX{{%
    \normalfont B\kern-0.5em{\scshape i\kern-0.25em b}\kern-0.8em\TeX}}}

\setcopyright{none} 
\renewcommand\footnotetextcopyrightpermission[1]{} 
\begin{document}

\title{A framework for spatial heat risk assessment using a generalized similarity measure} 

\author{Akshay Bansal}
\authornote{corresponding author}
\affiliation{%
  \department{Department of Computer Science}
  \institution{Virginia Tech}
  \country{Blacksburg, VA, United States}}
  \email{akshaybansal14@gmail.com}

\author{Ayda Kianmehr}
\affiliation{%
  \department{Spatial Sciences Institute}
  \institution{University of Southern California}
  \country{Los Angeles, CA, United States}}
\email{kianmehr@usc.edu}

\begin{abstract}

In this study, we develop a novel framework to assess health risks due to heat hazards across various localities (zip codes) across the state of Maryland with the help of two commonly used indicators: \textit{exposure} and \textit{vulnerability}. Our approach quantifies each of the two aforementioned indicators by developing their corresponding feature vectors and subsequently computes indicator-specific reference vectors that signify a high risk environment by clustering the data points at the tail-end of an \textit{empirical} risk spectrum. The proposed framework circumvents the information-theoretic entropy based aggregation methods whose usage varies with different views of entropy that are subjective in nature and more importantly generalizes the notion of risk-valuation using cosine similarity with unknown reference points.
  
\end{abstract}


\maketitle
\pagestyle{plain}
\section{Introduction}
Climate change is now one of the most well-known environmental phenomena of the 21st century. Within cities, severe climatic events, such as heatwaves, represent tangible outcomes of climate change, resulting in both direct and indirect impacts on the health and well-being of citizens. These effects encompass a spectrum from slight thermal discomfort due to increased temperatures to severe heat stress and, in the most extreme cases, heat-related fatalities \cite{theoharatos2010heat}. Between $2004$ and $2018$, the Center for Disease Control and Prevention (CDC) recorded an average of $702$ heat-related deaths per year across the United States \cite{cdc}.

Cities, due to their specific urbanization features, such as impervious surfaces, land cover, and morphological characteristics \cite{grimmond2007urbanization, kianmehr2022quantifying}, are more vulnerable to the impacts of such extreme events. Given that risk is determined by the interplay of \textit{hazard}, \textit{exposure}, and \textit{vulnerability} \cite{huang2020assessing}, more intensive risk will be associated with the extreme heat events throughout the cities. Therefore, the location-based risk identification and estimating \textit{relative} heat \textit{vulnerability} indices (HVI) scores are increasingly being applied by urban planners to map spatial distribution of heat risk and identify prioritized places for interventions and investments. However, upon the growing threat of heat waves, developing more accurate methods for risk assessment and quantification seems critical and necessary. 
\newline

\section{Related Research}
As it was noted by Intergovernmental Panel on Climate Change (IPCC) ($2014$), impacts from extreme climate-related events emerge from risk that are not only related to a specific hazard (e.g., heat waves), but also directly depends on the two other elements; \textit{exposure} and \textit{vulnerability}. Exposure addresses the population and assets at risk while \textit{vulnerability} indicates the susceptibility of human and natural systems during an extreme event\cite{pachauri2014climate}.

In general, socioeconomic profile of cities and neighborhoods can play a significant role for assessing the \textit{vulnerability} towards the risk of extreme heat. In a study in Phoenix and Philadelphia, using census data, it was demonstrated that heat mortality and distress are more rampant among vulnerable and marginalized groups. According to this study, most fatalities in these two cities during extreme heat events were mapped to Black, Hispanic, socially isolated and low value housing neighborhoods\cite{uejio2011intra}. In a similar study in Baltimore, Maryland it was shown that there is a strong correlation between surface temperature and socioeconomic factors. In the neighborhoods with low income, less educated and more ethnic minority residents, higher surface temperature were reported \cite{huang2011everyone}. This issue is mainly due to the fact that areas with underrepresented and marginalized residents are refrained from services such as tree planting and provision of park and open spaces.

More recently, significant efforts have been made to quantify the risk associated with extreme heat in cities using demographic data. These studies generally uses the three components of environmental risk, namely \textit{hazard}, \textit{exposure}, and \textit{vulnerability} to establish the heat risk index for spatial risk assessments. For example, in a study in China, historical meteorological data and socioeconomic statistics were used as indicators of risk components to evaluate the heat risk in $296$ cities and identify the spatial pattern of extreme heat risk \cite{huang2020assessing}. In a similar study, \cite{harlan2013neighborhood} census data were utilized to map vulnerable areas that were affected from extreme heat \textit{exposure} in Maricopa County, Arizona between $2000-2008$. This study also illustrates a strong correlation between heat mortality and socioeconomic characteristics of different neighborhoods. 

Although valuable efforts have been made so far for risk quantification, most of existing approaches of exposure-response lack the required accuracy for speculating the risk associated with heat \cite{mallen2019methodological}. This issue might emanate from the fact that those studies mainly consider different entropic measures which are mostly subjective in nature. Therefore, in this work, through using a generalized similarity measure, we attempt to develop a different approach to quantify two important elements of risk (i.e., \textit{exposure} and \textit{vulnerability}) and estimate the spatial distribution of risk associated with an extreme heat event. To apply our proposed approach of risk assessment in a real case, we chose Maryland, USA as our case study. Reasons for choosing this case study were the availability of demographic data and also the necessity of addressing extreme heat risk in this state (the prevalence of hot summer days and high heterogeneity in population's \textit{vulnerability} level). It is worth noting that in our analysis, we make an assumption that the heat event would almost uniformly affect all zip codes of Maryland as a result of which, the \textit{hazard} element of risk would be identical across all zip codes, and thus has not been considered in our calculation of heat risk.

\section{Methodology}
Next we discuss our approach along with its underlying theorems (and their corresponding proof sketch) to estimate the \textit{relative} heat-risk index for different zip codes across the state of Maryland. We will start with some basic assumptions and notations.

\subsection{Assumptions and notations}
Let $N$ be the total no. of zip codes across Maryland (as per the available data), then the \textit{exposure} and \textit{vulnerability} factors for a zip code indexed by $i \in \{0,1 \ldots N-1\}$ are encoded by the vectors $e_i$ and $v_i$ respectively. For our analysis, we assume that the extend of \textit{hazard} is same across all zip codes in the state. 

In addition, we assume the existence of reference vectors $e_*$ and $v_*$ that corresponds to the heat-risk of highest degree. Specifically, if the \textit{exposure} and \textit{vulnerability} of a certain zip code is encoded by the vectors $e_*$ and $v_*$, then the particular zip code has risk value of unity where in all our calculations the heat-risk index ($\mathcal{R}_i$) is scaled between $0$ and $1$ (i.e. $0 \leq \mathcal{R}_i \leq 1, \forall i \in [N]$). The risk index of $0$ implies minimal risk whereas the index value of $1$ signifies maximal \textit{relative} heat-risk situation across all zip codes of the state. 
\\

\textit{Notion of inner product: } If $p$ and $q$ are two real vectors of the same dimension, then the inner product $\inner{p}{q}$ is defined as $p^Tq$ where $p^T$ is the transpose of the vector $p$. Similarly, if $P$ and $Q$ are \textit{symmetric} matrices of the same dimension, then $\inner{P}{Q}$ is defined as $\tr(P^TQ) = \tr(PQ)$ where $\tr(.)$ is the sum of diagonal elements of the matrix under operation.

\subsection{Empirical risk estimation and data preprocessing}

We base our assumption of mutual independence of \textit{exposure} and \textit{vulnerability} in order to estimate heat-risk with the following formulation henceforth referred to as \textit{product-of-vector-inners} :
\begin{equation}\label{eq:risk:vector}
  \mathcal{R}_i = \left| \Bigl\langle {\frac{e_i}{\norm{e_i}}},{\frac{e_*}{\norm{e_*}}} \Bigr \rangle \right|. \left| \Bigl\langle{\frac{v_i}{\norm{v_i}}},{\frac{v_*}{\norm{v_*}}} \Bigr \rangle \right|.
\end{equation}

\textit{Validity of \textit{product-of-vector-inners}: } Evaluation of heat-risk using \cref{eq:risk:vector} satisfy most of the basic conditions for a valid estimate as the constituent inner products are non-negative and bounded above by unity ($\because$ Cauchy-Schwarz Inequality) implying $0 \leq \mathcal{R}_i \leq 1$. The heat-risk index evaluates to the maximum possible value of $1$ whenever $e_i = e_*$ and $v_i = v_*$. Also, for the reasons that would be evident later, a hypothetical case where $e_i$ and $v_i$ are zero vectors, the risk index should evaluate to $0$ that aligns with \cref{eq:risk:vector}.

It is worth noting that we need to appropriately scale the vectors $e_i$ and $v_i$ such that their components are all non-negative and intuitively, the lower values of the components should indicate a lower heat-risk index. In addition to the scaling objectives, we normalize the vectors $e_i$ and $v_i$ such that the resulting vectors all have unit $\ell_2$ norms. In a nutshell, we pre-process the data as per the following rules:
\begin{enumerate}\label{assumption:preprocessing}
  \item Transform the features to a non-negative scale such that lower values indicate lower risk.
  \item Scale down the value of the features below unity using \textit{min-max normalization} (i.e., $x \leftarrow (x - x_{min})/(x_{max} - x_{min})$).
  \item Normalize the vector by its $\ell_2$ norm (i.e., $v \leftarrow v/\norm{v}_2$)
\end{enumerate}

With respect to risk estimate discussed in \cref{eq:risk:vector}, we observe the following equivalence via the lemma:
\begin{lemma}
  The empirical risk estimate given by \cref{eq:risk:vector} is equivalent to:
  \begin{equation}\label{eq:risk:vec:transpose}
    \mathcal{R}_i = \sqrt{\Bigl\langle {\frac{e_i e_i^T}{\norm{e_i}_2^2}},{\frac{e_*e_*^T}{\norm{e_*}_2^2}} \Bigr \rangle}. \sqrt{\Bigl\langle{\frac{v_iv_i^T}{\norm{v_i}_2^2}},{\frac{v_*v_*^T}{\norm{v_*}_2^2}} \Bigr \rangle}.
  \end{equation}
\end{lemma}

\begin{proof}
  (Sketch) Trivially follows from the symmetric and cyclic properties of trace operator which is a valid inner product.
\end{proof}

\subsection{Generalized empirical risk estimate}
Next we consider generalizing \cref{eq:risk:vec:transpose} to develop the risk estimate in the following form:
\begin{equation}\label{eq:risk:mat}
  \mathcal{R}_i = \sqrt{\inner{e_ie_i^T}{E_*}}.\sqrt{\inner{v_iv_i^T}{V_*}}
\end{equation}
with the additional constraints that the reference objects for \textit{exposure} and \textit{vulnerability}, $E_*$ and $V_*$ are symmetric positive semi-definite matrices (smallest eigenvalue being non-negative) where $\tr(E_*) = \tr(V_*) = 1$. Note that unlike the rank-one reference matrices depicted in \cref{eq:risk:vec:transpose}, the reference matrices in \cref{eq:risk:mat} can possibly be of rank greater than unity. The vectors $e_i, v_i$ are assumed to be normalized post pre-processing (according to \cref{assumption:preprocessing}).

In order to come up with a good estimate of the reference objects $E_*$ and $V_*$, we devise heuristic vectors $e_H$ and $v_H$ described in the following way:
\begin{align}
  e_H^{(j)} &= 1 / \sqrt{L_E} \,\, (\forall j) \label{eq:heuristic:vec:exp}\\
  v_H^{(j)} &= 1 / \sqrt{L_V} \,\, (\forall j) \label{eq:heuristic:vec:vul}
\end{align}
where the superscript $j$ denotes the $j$-th component of the underlying vector while $L_E$ and $L_V$ corresponds to the length of the \textit{exposure} and \textit{vulnerability} vectors respectively. With the help of the computed heuristic vectors, we subsequently construct the sets:
\begin{align}
  \mathcal{S}^{(E)} &= \{\inner{e_i}{e_H} : i \in [N]\} \label{eq:heuristic:set:exp}\\
  \mathcal{S}^{(V)} &= \{\inner{v_i}{v_H} : i \in [N]\}. \label{eq:heuristic:set:vul}
\end{align}
We cluster the points on the right tail-end of the distributions in the above sets to form another pair of tail-valued subsets $\mathcal{S}^{(E)}_{T} \subset \mathcal{S}^{(E)}$ and $\mathcal{S}^{(V)}_{T} \subset \mathcal{S}^{(V)} $. Our method for evaluating heuristic vectors using equations \ref{eq:heuristic:vec:exp},\ref{eq:heuristic:vec:vul} and eventually zeroing down on the subset using the points on the tail-end distribution by constructions \ref{eq:heuristic:set:exp},\ref{eq:heuristic:set:vul} points to the fact that the vectors contained in the sets $\mathcal{S}^{(E)}_T$ and $\mathcal{S}^{(V)}_T$ are \textit{conditional} high risk states.
In the next subsection, we will develop a technique to find matrix object(s) that can serve as a measure for central tendency for the two sets $\mathcal{S}^{(E)}_T$ and $\mathcal{S}^{(V)}_T$ satisfying the generalized estimate conditions of being \textit{positive semi-definite} with unit trace as discussed previously in the necessary conditions for \cref{eq:risk:mat}.

\subsection{Distance measure for semi-definite states}
Let $P$ and $Q$ be two symmetric positive \textit{semi-definite} matrices with unit trace, then a measure to gauge similarity (or distance) between $P$ and $Q$ is given by \fidelity. The rationale behind using \fidelity as a distance measure roots in its usage in quantum information theory for quantum mixed states which belong to the convex cone of positive semi-definite matrices defined over general \textit{Hilbert} spaces \cite{nielsen2002quantum}. Mathematically, we define the \fidelity measure between matrices $P$ and $Q$:
\begin{equation}\label{eq:fidelity}
  F(P,Q) = \tr\sqrt{P^{\frac{1}{2}}QP^{\frac{1}{2}}}.
\end{equation}

\begin{lemma}\label{fidelity:prop}
   The \fidelity measure defined in \cref{eq:fidelity} commutes and is symmetric in $P$ and $Q$ with $0 \leq F(P,Q) \leq 1$.
\end{lemma}

From the definition above, we can establish an equivalent representation for \cref{eq:risk:mat} in the following form:
\begin{equation}\label{eq:risk:fidelity}
  \mathcal{R}_i = F(e_ie_i^T, E_*).F(v_iv_i^T, V_*).
\end{equation}

Due to the unavailability of reference states $E_*$ and $V_*$, we develop their estimates $\hat{E}_*$ and $\hat{V}_*$ whose calculation is described next.

\begin{lemma}
  Let,
  \begin{align}
    \hat{E}_* &= \textup{argmax}_{\rho} \sum_{e \in \mathcal{S}^{(E)}_{T}}F(ee^T,\rho) : \rho \succcurlyeq 0, \tr(\rho) = 1 \label{eq:estimate:E}\\
    \hat{V}_* &= \textup{argmax}_{\sigma} \sum_{v \in \mathcal{S}^{(V)}_{T}}F(vv^T,\sigma) : \sigma \succcurlyeq 0, \tr(\sigma) = 1 \label{eq:estimate:V}
  \end{align}
  then $\hat{E}_* $ and $\hat{V}_*$ are valid estimates for $E_*$ and $V_*$ respectively.
\end{lemma}
\begin{proof}
  (Sketch) The constraints on $\rho$ and $\sigma$ being positive semi-definite matrices with unit trace in \cref{eq:estimate:E} and \cref{eq:estimate:V}, and the objective functions that maximizes the aggregate similarity between points imply that the optimal $\rho$ and $\sigma$ are valid measures of central tendency for clusters $\mathcal{S}^{(E)}_{T}$ and $\mathcal{S}^{(V)}_{T}$ respectively.
\end{proof}
The optimization problems defined in \ref{eq:estimate:E} and \ref{eq:estimate:V} could be solved efficiently if these are expressed as convex optimization problems (convex objective with convex constraints). In order to achieve the same, we use the following theorem:

\begin{theorem}\label{fidelity:convexity}
  The \fidelity function $F(\rho, P)$ for a fixed $P \in \mathbb{R}^{q \times q}$ is a concave function and can be expressed as the following semi-definite program:
  \begin{equation}
    F(\rho, P) =  \sup \Bigl\langle \begin{bmatrix}
      0 & {I}/2 \\
      {I}/2 & 0
    \end{bmatrix}, X \Bigl\rangle
  \end{equation}
subject to:
\begin{align*}
  X &= \begin{bmatrix}
    \rho & R \\
    S & P
  \end{bmatrix}, \\
  X &\succcurlyeq 0, \\
  X &\in \mathbb{R}^{2q \times 2q}, \\
  R,S &\in \mathbb{R}^{q \times q}.
\end{align*}
\end{theorem}
\begin{proof}
  (Sketch) The concave nature of the function $F(\rho,P)$ could be shown by parameterizing $\rho$ on fixed line and evaluating the double differential to return a negative hessian over the defined parameter.  
\end{proof}
With the help of \cref{fidelity:convexity}, we now prove the convexity of the problems \ref{eq:estimate:E} and \ref{eq:estimate:V}:
\begin{theorem}\label{theorem:sdp:estimate}
  The problem \ref{eq:estimate:E} (or \ref{eq:estimate:V}) is convex with its semi-definite formulation given by:
  \begin{equation}
    \sup_{\{X_j\}} \sum_{j \in [\mathcal{S}^{(E)}_{T}] : e_{t_j} \in \mathcal{S}^{(E)}_{T} } \Bigl\langle \begin{bmatrix}
      0 & {I}/2 \\
      {I}/2 & 0
    \end{bmatrix}, X_j \Bigl\rangle
  \end{equation}
subject to:
\begin{align*}
  X_j &= \begin{bmatrix}
    \rho & R_j \\
    S_j & e_{t_j}e_{t_j}^T
  \end{bmatrix}, \\
  X_j &\succcurlyeq 0,  \\
  X_j &\in \mathbb{R}^{2q \times 2q}, \\
  R_j,S_j &\in \mathbb{R}^{q \times q}
\end{align*}
where $e_i \in \mathbb{R}_{+}^{q}$.
\end{theorem}

\begin{proof}
  (Sketch) The concave nature of the objective function in problem \ref{eq:estimate:E} is implied by the fact that it is a sum of concave functions with extended linear matrix inequality constraints \cite{boyd2004convex}. Additionally, the sum of maximum inner products would be equal to the maximum of the sum of inner products as the constraints in the formulation are non-overlapping.   
\end{proof}

\subsection{Summary of the approach}
Finally, we summarize the process of estimating heat-risk index for our work:
\begin{enumerate}
  \item Pre-process the data vectors $e_i, v_i$ as described in \cref{assumption:preprocessing}.
  \item Calculate the heuristic vectors $e_H, v_H$ using \cref{eq:heuristic:vec:exp} and \cref{eq:heuristic:vec:vul}.
  \item Construct the subsets $\mathcal{S}^{(E)}_{T}, \mathcal{S}^{(V)}_{T}$ from $\mathcal{S}^{(E)}, \mathcal{S}^{(V)}$ (given by \ref{eq:heuristic:set:exp}, \ref{eq:heuristic:set:vul}).
  \item Calculate the estimates $\hat{E}_*, \hat{V}_*$ by formulating problems \ref{eq:estimate:E}, \ref{eq:estimate:V} as semi-definite optimization problems as described in \cref{theorem:sdp:estimate}.
  \item Compute the \textit{relative} heat-risk index for all zip codes with the available data using \cref{eq:risk:mat} where the reference matrices $E_*, V_*$ are substituted by their estimates $\hat{E}_*, \hat{V}_*$ calculated in the previous step.
\end{enumerate}

\section{Experiments}
  \subsection{Data}
We apply our proposed approach for Maryland to evaluate the distributed heat-related for a set of $467$ zip codes across the state. Using Maryland's most recent census data \href{https://data.imap.maryland.gov/datasets/eb706b48117b43d482c63d02017fc3ff/explore?location=38.813879}{(link)}, published by United States Census Bureau for these $467$ zip codes, we derive relevant criteria associated with \textit{exposure} and \textit{vulnerability} elements. List of variables which were used as the indicators of \textit{exposure} and \textit{vulnerability} are represented in \cref{table:variables:indicators}.

\begin{table}[]
\caption{Considered variables for experiment}`
\centering
\label{table:variables:indicators}
\begin{tabular}{|l|l|l|l|}
\hline
\textbf{Variable} &
  \textbf{Description} &
  \textbf{Indicator} &
  \textbf{Risk element} \\ \hline
\begin{tabular}[c]{@{}l@{}}Population\\ density\end{tabular} &
  \begin{tabular}[c]{@{}l@{}}Total \\ population \\ to the\\ total area\end{tabular} &
  \begin{tabular}[c]{@{}l@{}}Exposed \\ population\end{tabular} &
  Exposure \\ \hline
\begin{tabular}[c]{@{}l@{}}Housing\\ density\end{tabular} &
  \begin{tabular}[c]{@{}l@{}}Total\\ housing units \\ to\\ the total area\end{tabular} &
  \begin{tabular}[c]{@{}l@{}}Exposed \\ residents\end{tabular} &
  Exposure \\ \hline
\begin{tabular}[c]{@{}l@{}}Black\\ population\\rate\end{tabular} &
  \begin{tabular}[c]{@{}l@{}}Black \\ population to \\ the overall \\ population\end{tabular} &
  \begin{tabular}[c]{@{}l@{}}Race/\\ ethnicity\end{tabular} &
  vulnerability \\ \hline
\begin{tabular}[c]{@{}l@{}}Hispanic\\ population\\rate\end{tabular} &
  \begin{tabular}[c]{@{}l@{}}Hispanic \\ population to \\ the \\ overall \\ population\end{tabular} &
  \begin{tabular}[c]{@{}l@{}}Race/\\ ethnicity\end{tabular} &
  vulnerability \\ \hline
\begin{tabular}[c]{@{}l@{}}American \\ Indian \\ population\\rate\end{tabular} &
  \begin{tabular}[c]{@{}l@{}}American \\ Indian \\ population \\ to the \\ overall \\ population\end{tabular} &
  \begin{tabular}[c]{@{}l@{}}Race/\\ ethnicity\end{tabular} &
  vulnerability \\ \hline
\begin{tabular}[c]{@{}l@{}}Native \\ Hawaiian\\ population\\rate\end{tabular} &
  \begin{tabular}[c]{@{}l@{}}Native \\ Hawaiian \\ population \\ to the overall \\ population\end{tabular} &
  \begin{tabular}[c]{@{}l@{}}Race/\\ ethnicity\end{tabular} &
  vulnerability \\ \hline
\begin{tabular}[c]{@{}l@{}}Population \\ over\\ age 65\\rate\end{tabular} &
  \begin{tabular}[c]{@{}l@{}}Elderly \\ population \\ (over age 65) \\ to the overall \\ population\end{tabular} &
  \begin{tabular}[c]{@{}l@{}}Age/\\ well being\end{tabular} &
  vulnerability \\ \hline
\begin{tabular}[c]{@{}l@{}}Population \\ below\\ age 18\\rate\end{tabular} &
  \begin{tabular}[c]{@{}l@{}}Young \\ population \\ (below age 18) \\ to the overall \\ population\end{tabular} &
  \begin{tabular}[c]{@{}l@{}}Age/\\ well being\end{tabular} &
  vulnerability \\ \hline
\begin{tabular}[c]{@{}l@{}}Rental \\ housing\\ rate\end{tabular} &
  \begin{tabular}[c]{@{}l@{}}Rental \\ housing to the \\ overall \\ housing units\end{tabular} &
  \begin{tabular}[c]{@{}l@{}}Income/\\ economic \\ status\end{tabular} &
  vulnerability \\ \hline
\end{tabular}
\end{table}

Variables of \cref{table:variables:indicators} are chosen based on the \textit{exposure} and \textit{vulnerability} indices reported in the literature. Racial disparities in heat-related mortalities during heatwaves (e.g., Chicago heatwave in $1995$) strengthened the hypothesis that heat related fatalities among some racial and ethnic groups (such as Black and Hispanic) are more rampant than others. Some references associated this fact to the genetic differences \cite{hansen2013vulnerability}. However, this racial difference in heat tolerance can also be attributed to the living situations, lower income, poorer environmental conditions such as sparse vegetation and lower air conditioning ownership \cite{harlan2006neighborhood}\cite{o2005disparities}.

In addition, social isolation and linguistic isolation might play a role in the incapability of minority racial groups to understand heat warnings and thus their failure to adopt suggested principals by government and escape to the provided cooling centers during heatwaves \cite{uejio2011intra}. This issue can be further exacerbated by their probable concerns for the immigration status and deportation \cite{richard2011correlates}. 

Failing to comprehend the official educational messages and taking appropriate actions during heat wave is not uncommon for elderly population as well. Additionally, physical health issues, chronic diseases and disabilities can make elderly population more vulnerable to the risk of extreme heat \cite{aastrom2011heat}\cite{yu2012daily}. Similar reasons owing to the dependency for travel and movement, also classify young population (below age $18$) to be a vulnerable set \cite{xu2014impact}. 

\begin{figure}[!htb]
\includegraphics[width=4cm, height=3cm]{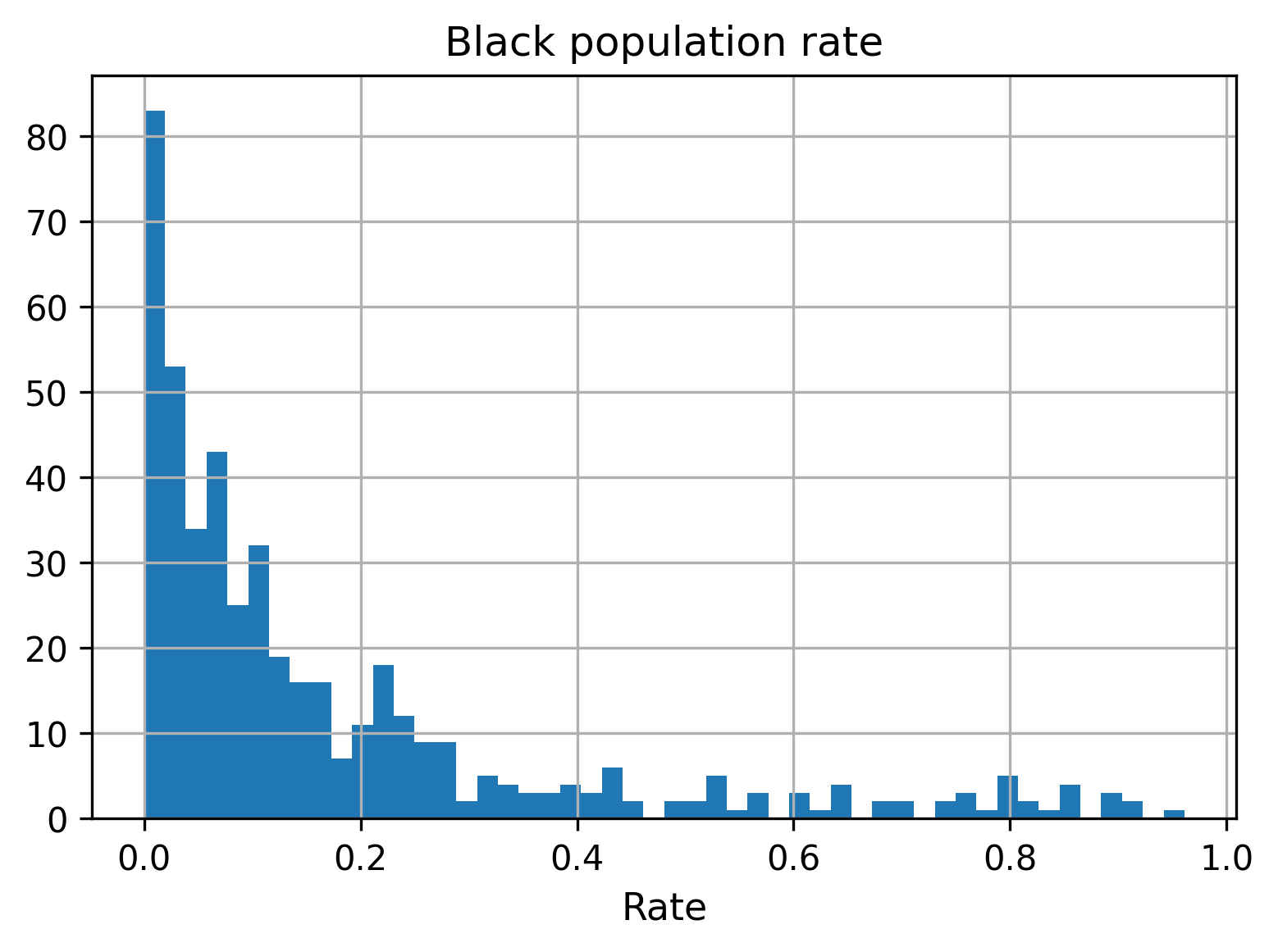}
\centering
\includegraphics[width=4cm, height=3cm]{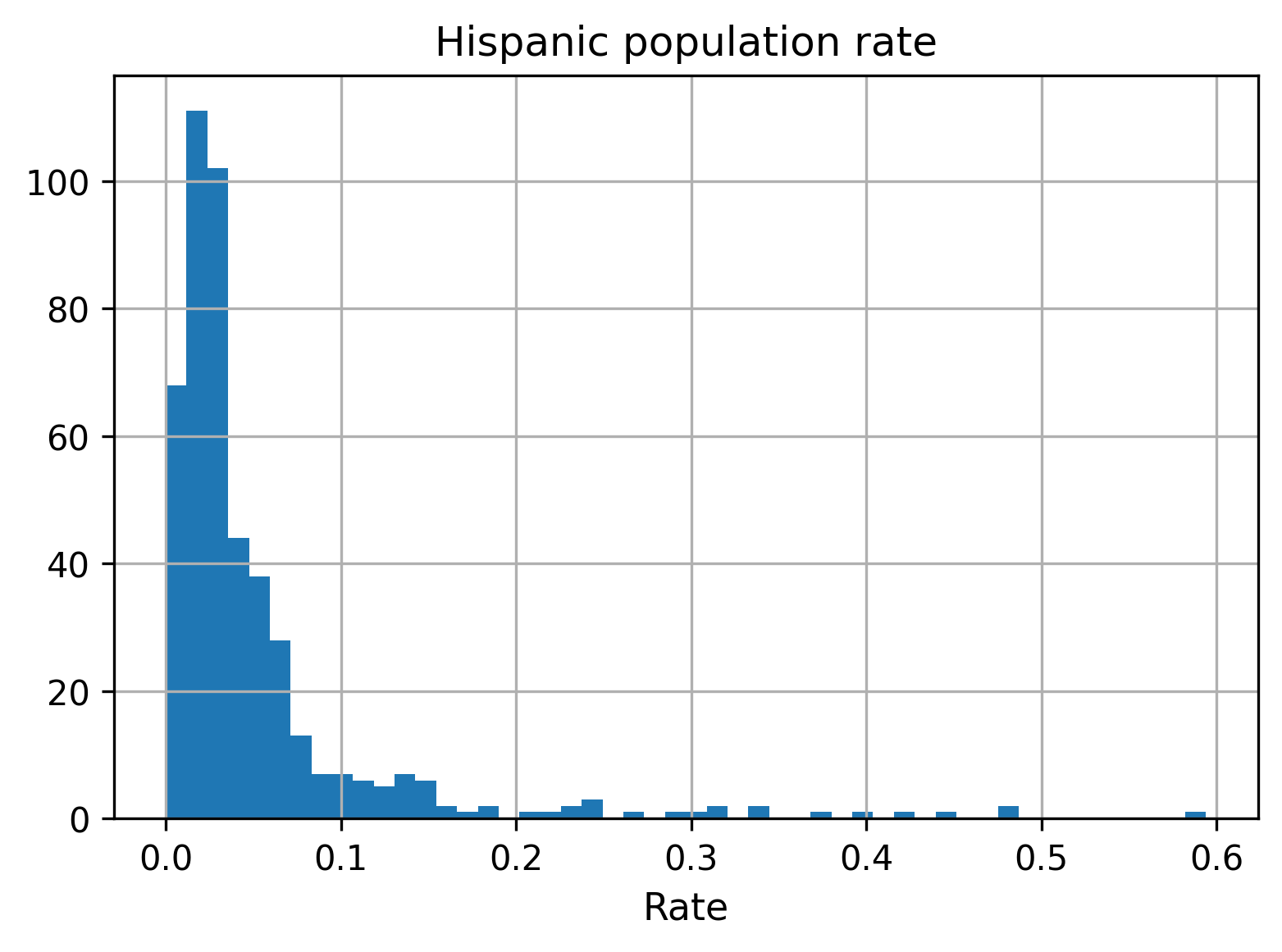}
\newline
\includegraphics[width=4cm, height=3cm]{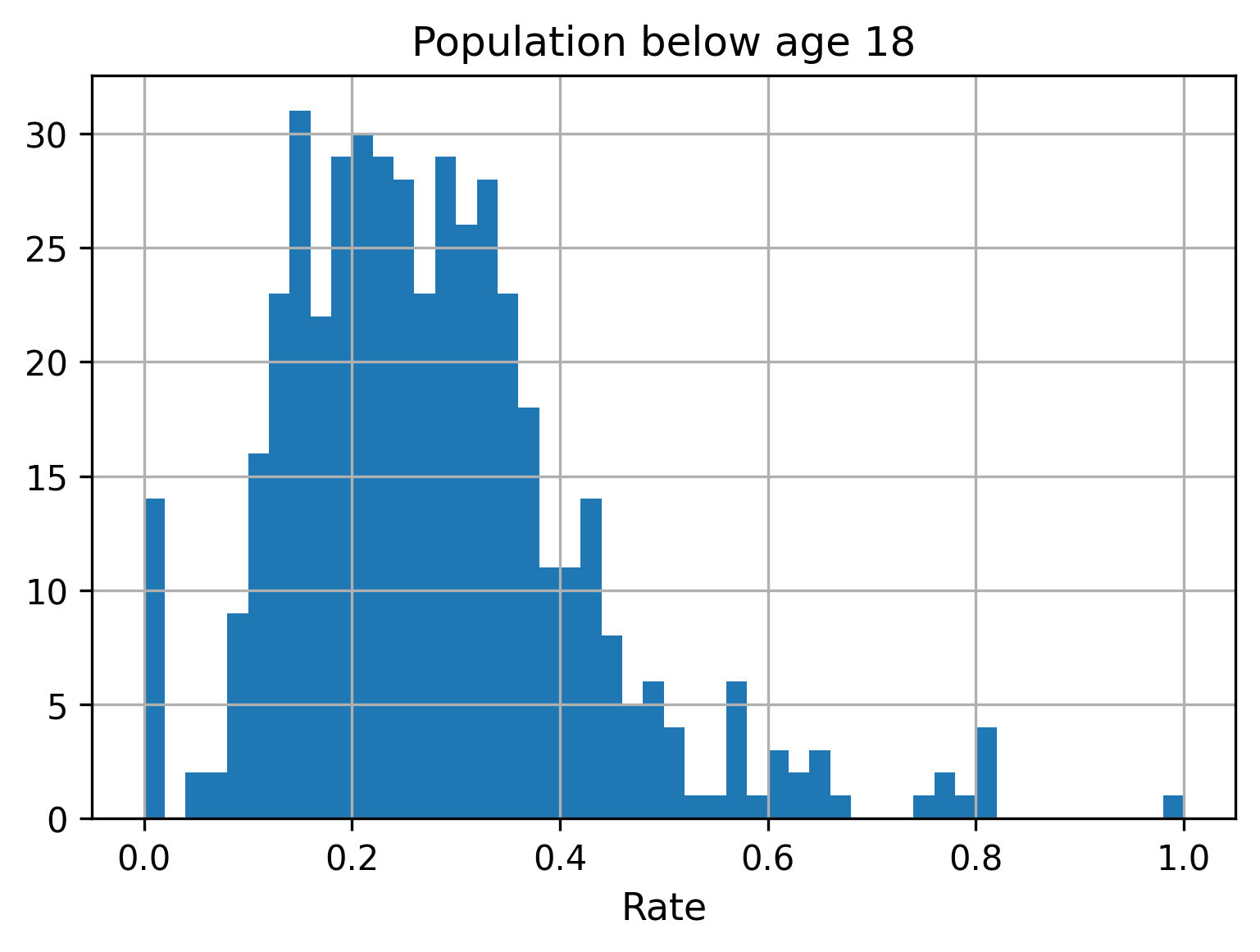}
\centering
\includegraphics[width=4cm, height=3cm]{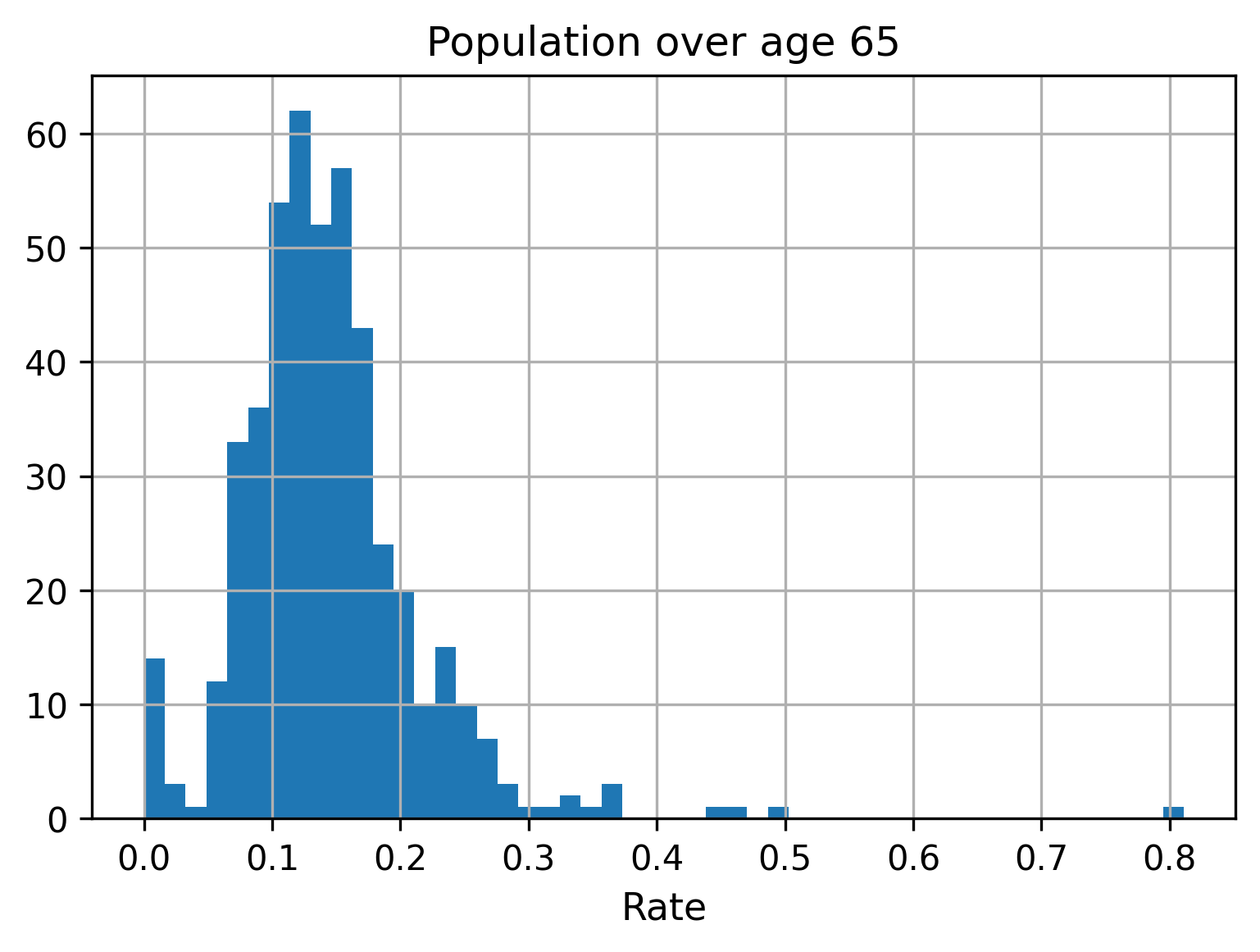}
\includegraphics[width=4cm, height=3.19cm]{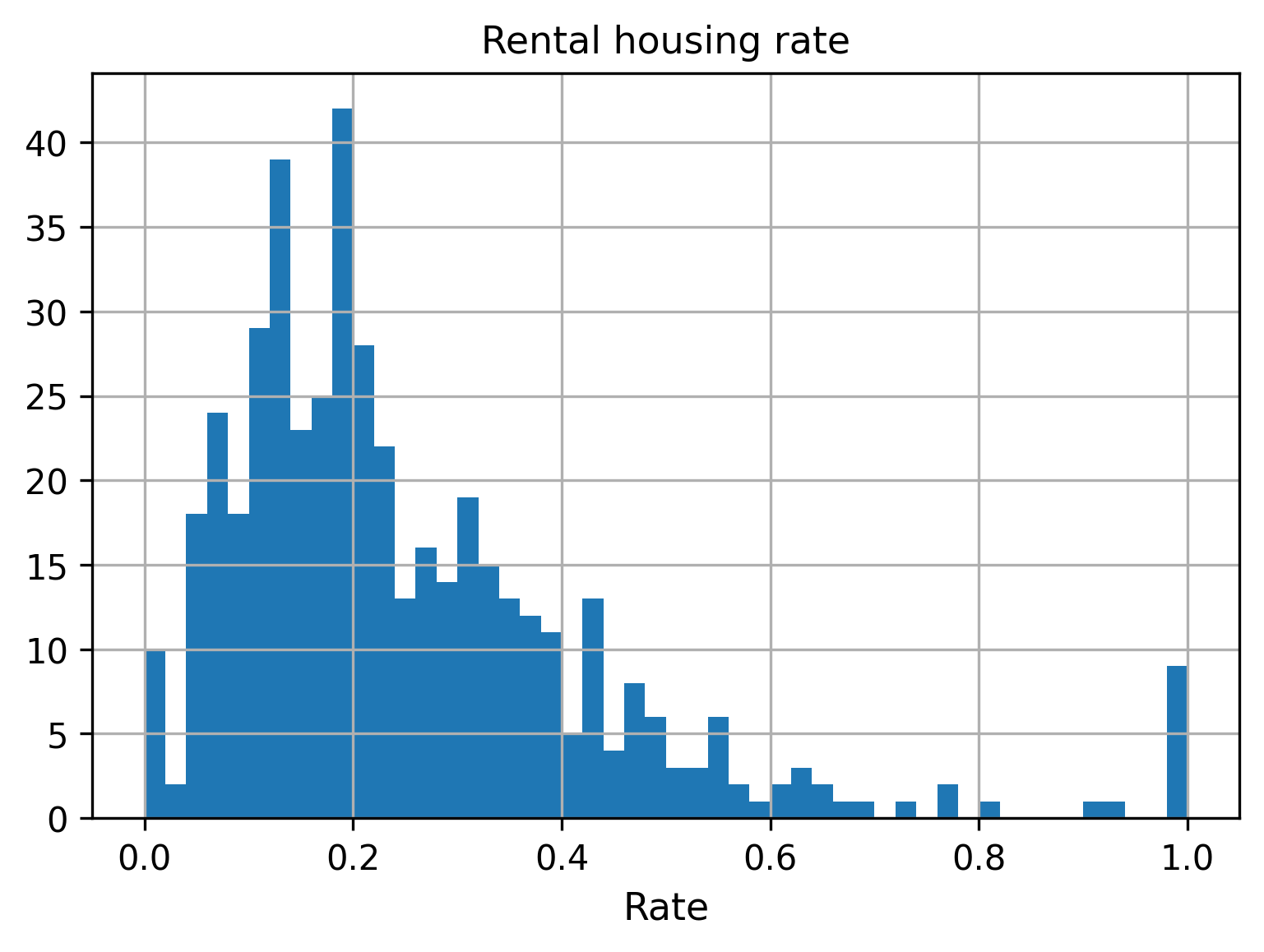}
\centering
\includegraphics[width=4cm, height=3.2cm]{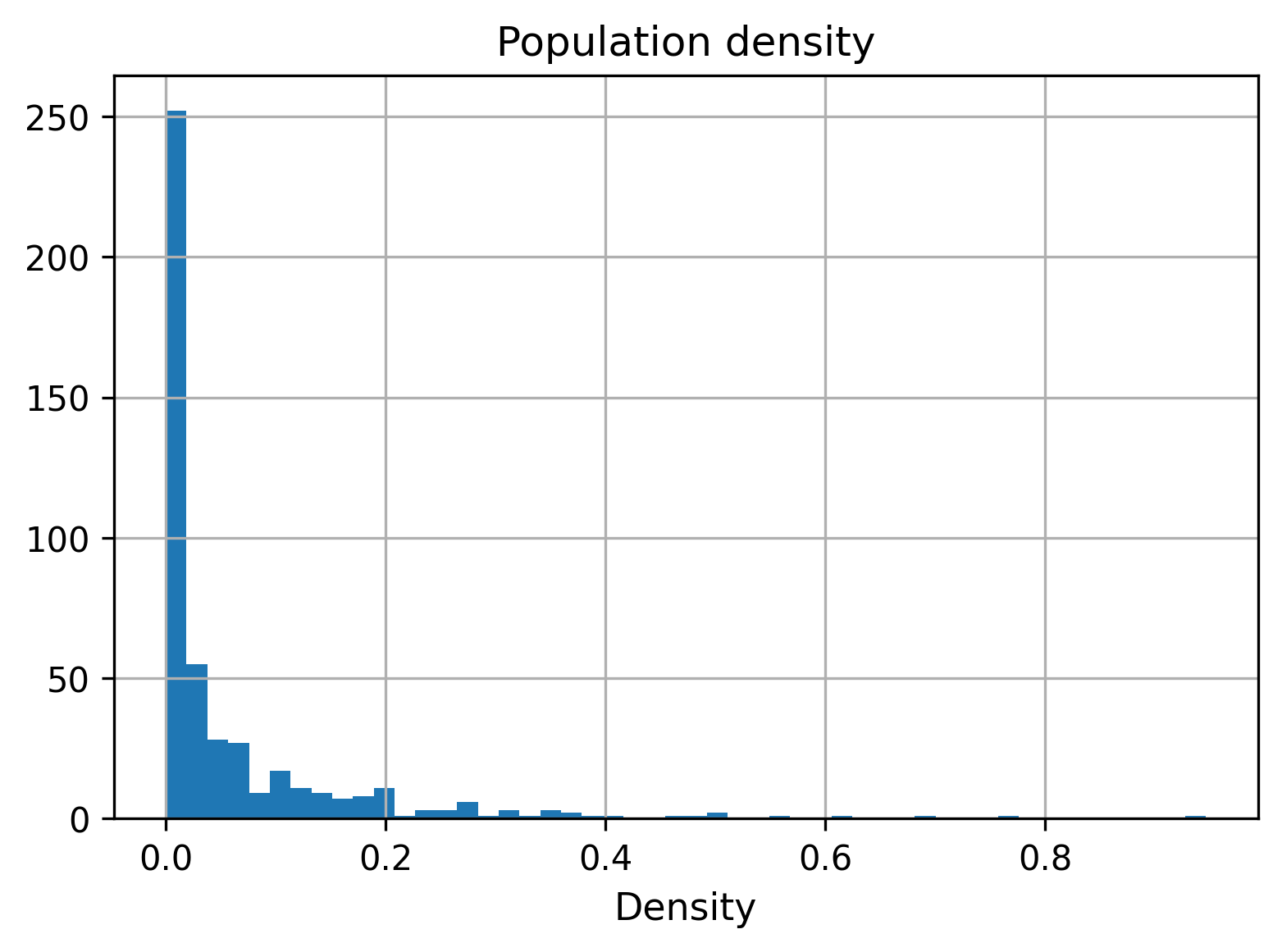}
\caption{Histograms of selected risk variables}
\label{fig:histogram:risk:variables}
\end{figure}

Based on the above facts, we also consider race / ethnicity, age / well-being, income and economic status as the indicators of \textit{vulnerability} to extreme heat. Ergo, we chose variables such as fraction of Black, Hispanic, Native Hawaiian, American Indian, population over the age of $65$ and below the age of $18$ separately to represent the vulnerabilities from Maryland census data. Exploratory analysis of four important \textit{vulnerability} indicators (Black / Hispanic population rate and population over age $65$ / below age $18$ presented in \cref{fig:histogram:risk:variables} suggests that age and well-being might be crucial for accurate risk prediction.

Moreover, as the higher population density and higher resident population rate in an area would make more population expose to the risk \cite{dong2020heatwave}, we chose the population density and the ratio of housing units in each zip code as the representative of \textit{exposure} element to quantify the risk associated with extreme heat.

  \subsection{Setup}
  We applied our method for \textit{relative} heat-risk calculation by creating vectors that represents \textit{exposure} and \textit{vulnerability} across different zip codes for the state of Maryland. The features for heat \textit{exposure} included population and housing density while the \textit{vulnerability} was primarily represented by an aggregation of heat-sensitive ethnic groups (Hispanic and non-Hispanic minorities), age-groups and percentage of population with rented housing.
  
\begin{figure}[!htb]
    \centering
    \includegraphics[width=8cm, height=4cm]{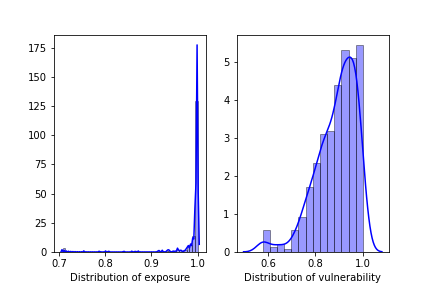}
    \caption{Distribution of indicators w.r.t. heuristic reference}
    \label{fig:distribution:indicator}
 \end{figure}

  Due to the left-skewed distribution of inner products for \textit{exposure} specific data vectors (\cref{fig:distribution:indicator}), the set $\mathcal{S}^{(E)}_T$ (\cref{eq:heuristic:set:exp}) was constructed by collecting data vectors with similarity greater than the estimated mean while the set $\mathcal{S}^{(V)}_T$ (\cref{eq:heuristic:set:vul}) was formed by collecting vectors with similarity greater than estimated mean added to its standard deviation (\cref{fig:distribution:indicator}). The cluster centers or the high-rank approximation matrices for each of the tailed-sets was calculated by solving the semi-definite program defined in \cref{theorem:sdp:estimate} using an SDP solver \href{https://www.mosek.com/documentation/}{MOSEK} to report the risk-indices as devised in \cref{eq:risk:mat}.
  
  \subsection{Results}
    The resultant optimal estimates of reference matrices ($E_*$ and $V_*$)  for high \textit{exposure} and \textit{vulnerability} are shown next:
    \begin{equation}
        \hat{E}_* = \begin{bmatrix}
            0.51578382 & 0.49975081 \\
            0.49975081 & 0.48421618
            \end{bmatrix} 
    \end{equation}
    \begin{equation}
        \hat{V}_* = \begin{bmatrix}
            0.34086951 & 0.35583112 & 0.31314804 \\
            0.35583112 & 0.37144944 & 0.3268929 \\
            0.31314804 & 0.3268929  & 0.28768105
        \end{bmatrix}
    \end{equation}
    The principle eigenvector of $\hat{V}_* $ i.e. $\begin{bmatrix} 0.58384031 \\ 0.80354261 \\  0.11597316 \end{bmatrix}^T$ is notably different from the heuristic reference vector of $\begin{bmatrix} 0.577350 \\ 0.577350 \\ 0.577350 \end{bmatrix}^T$ implying that appropriate reference is crucial to evaluate \textit{relative} heat-risk index. The estimated heat-risk indices using this approach corroborates the hypothesis that the zip codes with near uniform high concentration across all its variables (with properties defined in \cref{assumption:preprocessing}) tends to belong to the high-risk zone while the the zip codes with uneven concentration are comparatively risk averse.

    The final risk quantification product is depicted in fig. 3. As apparent in this map, the northern and west-northern parts of Maryland (mainly occupied by white-affluent residents) are relatively less susceptible to the risk associated with an extreme heat event, while the southern and central parts, which are inhabited more by racial minority groups, are at greater risk. Some zip codes due to the lower population density (and thus lesser exposure to extreme heat), were also identified as low risk zones.
    
 \begin{figure*}
  \includegraphics[width=15cm,height=7.5cm]{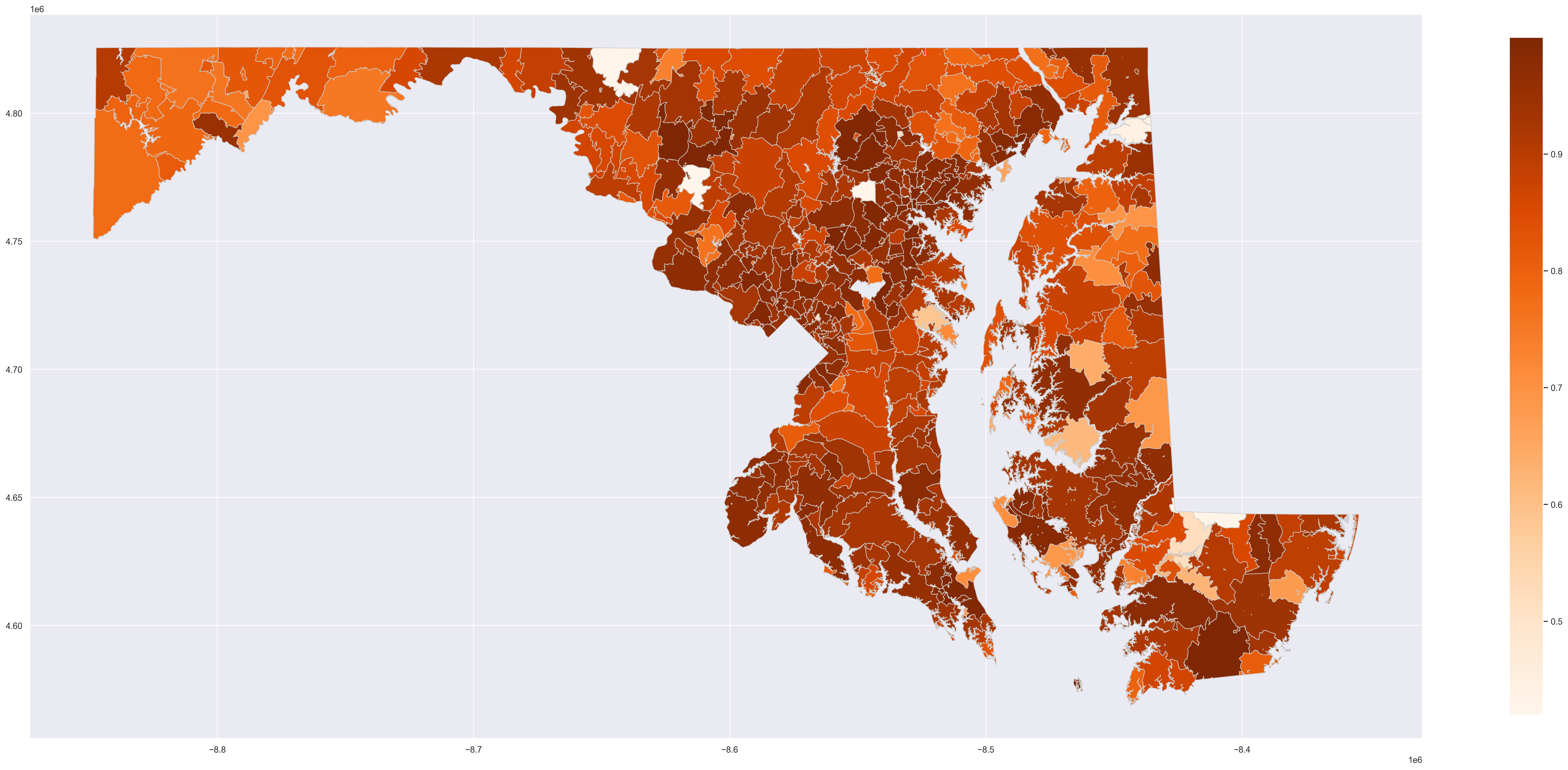}
  \caption{Estimated risk associated with extreme heat in each zip code in Maryland}
\end{figure*}

\section{Discussion}
    In this study, we primarily emphasized on the importance of calculating heat-risk index with a \textit{relative} rather an \textit{absolute} notion. The risk-estimates thus calculated can help devise appropriate actions at various structural or administrative levels to mitigate the effects of extreme heat hazards. In the process we noticed that highly concentrated vectors across all the spanning variables may not always be the best choice to set the reference and should be governed by a general central measure of the clusters of different high-risk indicators (such as \textit{exposure}, \textit{vulnerability}, \textit{hazard}).

    Although in our case, the lower order eigenvalues for $\hat{E}_*$ and $\hat{V}_*$ both tended to zero, but a general scenario where the vectors in the high-risk clusters are distributed heteroscedastically may yield reference objects of rank strictly greater than unity. 
  
    As previous studies demonstrated, other heat metrics, such as surface temperature and mean radiant temperature, exhibit high spatial variability and are correlated with social factors \cite{huang2011everyone, kianmehr2023comparison}. This implies that using such heat metrics might be relevant to improve the accuracy of our model for spatial heat risk assessment. 
    
    Moreover, with further information on the degree to which each variable might contribute to heat-related fatalities, more complex algorithms, including the differential weighting of various indicators, can be integrated into our framework to estimate the risk more accurately. 
    
    Our analysis in this paper was limited by the available data, so many contributing factors to the \textit{vulnerability} and the associated risk (such as housing qualities and materials, air conditioner availability, the green coverage ratio, education level, and social connections among individuals) were not considered in our analysis. Such information could be made available only through local surveys, which are usually expensive and time-consuming. Thus, the variables used in this study worked as a proxy for the other important factors mentioned previously.

\section{Conclusion}
This study, by proposing a new framework, tried to quantify the risk associated with heat in a more robust and objective manner. This risk-assessment framework can be utilized in cities with elevated risk of heat-related fatalities and would help planners and decision-makers take informed decisions, devise appropriate strategies and implement place-based interventions. 

As some of the past works in this direction included another indicator of \textit{hazard} for their analysis, in the future, we can possibly expand our work to landscapes with non-negligible spatial gradients due to heat hazard when conditioned on the other indicators of \textit{exposure} and \textit{vulnerability}. Furthermore, we can validate our proposed method of heat-risk mapping by comparing the risk indices with heat-related morbidity observations in a \textit{relative} sense.


\section{Acknowledgements}
We would like to thank Disaster Resiliency and Risk Management (DRRM) program at Virginia Tech for providing a context to thrive interdisciplinary thinking on the hazard and risk management and  Dr. Jamie Sikora (Dept. of Computer Science, Virginia Tech) for fruitful discussions on the intuition behind the usage of \textit{fidelity} measures.



The data used for this work was downloaded from \href{https://data.imap.maryland.gov/datasets/eb706b48117b43d482c63d02017fc3ff/explore?location=74.191427%2C-108.375741%2C4.12}{\textit{Maryland's Mapping \& GIS  Data Portal}}. The code developed for the analysis can be accessed at the git repository (\href{https://bitbucket.org/akshaybansal14/urban-heat-risk-assessment/src/master/}{link}).

\bibliographystyle{ACM-Reference-Format}
\bibliography{references}


\begin{thebibliography}{21}


\ifx \showCODEN    \undefined \def \showCODEN     #1{\unskip}     \fi
\ifx \showDOI      \undefined \def \showDOI       #1{#1}\fi
\ifx \showISBNx    \undefined \def \showISBNx     #1{\unskip}     \fi
\ifx \showISBNxiii \undefined \def \showISBNxiii  #1{\unskip}     \fi
\ifx \showISSN     \undefined \def \showISSN      #1{\unskip}     \fi
\ifx \showLCCN     \undefined \def \showLCCN      #1{\unskip}     \fi
\ifx \shownote     \undefined \def \shownote      #1{#1}          \fi
\ifx \showarticletitle \undefined \def \showarticletitle #1{#1}   \fi
\ifx \showURL      \undefined \def \showURL       {\relax}        \fi
\providecommand\bibfield[2]{#2}
\providecommand\bibinfo[2]{#2}
\providecommand\natexlab[1]{#1}
\providecommand\showeprint[2][]{arXiv:#2}

\bibitem[\protect\citeauthoryear{{\AA}str{\"o}m, Bertil, and
  Joacim}{{\AA}str{\"o}m et~al\mbox{.}}{2011}]%
        {aastrom2011heat}
\bibfield{author}{\bibinfo{person}{Daniel~Oudin {\AA}str{\"o}m},
  \bibinfo{person}{Forsberg Bertil}, {and} \bibinfo{person}{Rockl{\"o}v
  Joacim}.} \bibinfo{year}{2011}\natexlab{}.
\newblock \showarticletitle{Heat wave impact on morbidity and mortality in the
  elderly population: a review of recent studies}.
\newblock \bibinfo{journal}{\emph{Maturitas}} \bibinfo{volume}{69},
  \bibinfo{number}{2} (\bibinfo{year}{2011}), \bibinfo{pages}{99--105}.
\newblock


\bibitem[\protect\citeauthoryear{Boyd and Vandenberghe}{Boyd and
  Vandenberghe}{2004}]%
        {boyd2004convex}
\bibfield{author}{\bibinfo{person}{Stephen~P Boyd} {and}
  \bibinfo{person}{Lieven Vandenberghe}.} \bibinfo{year}{2004}\natexlab{}.
\newblock \bibinfo{booktitle}{\emph{Convex optimization}}.
\newblock \bibinfo{publisher}{Cambridge university press}.
\newblock


\bibitem[\protect\citeauthoryear{Dong, Peng, He, Corcoran, Qiu, and Wang}{Dong
  et~al\mbox{.}}{2020}]%
        {dong2020heatwave}
\bibfield{author}{\bibinfo{person}{Jianquan Dong}, \bibinfo{person}{Jian Peng},
  \bibinfo{person}{Xiaorong He}, \bibinfo{person}{Jonathan Corcoran},
  \bibinfo{person}{Sijing Qiu}, {and} \bibinfo{person}{Xiaoyu Wang}.}
  \bibinfo{year}{2020}\natexlab{}.
\newblock \showarticletitle{Heatwave-induced human health risk assessment in
  megacities based on heat stress-social vulnerability-human exposure
  framework}.
\newblock \bibinfo{journal}{\emph{Landscape and Urban Planning}}
  \bibinfo{volume}{203} (\bibinfo{year}{2020}), \bibinfo{pages}{103907}.
\newblock


\bibitem[\protect\citeauthoryear{for Disease~Control and Prevention}{for
  Disease~Control and Prevention}{2020}]%
        {cdc}
\bibfield{author}{\bibinfo{person}{Centre for Disease~Control} {and}
  \bibinfo{person}{Prevention}.} \bibinfo{year}{2020}\natexlab{}.
\newblock \bibinfo{booktitle}{\emph{Heat-Related Deaths — United States,
  2004–2018}}.
\newblock
\urldef\tempurl%
\url{https://www.cdc.gov/mmwr/volumes/69/wr/mm6924a1.htm}
\showURL{%
\tempurl}


\bibitem[\protect\citeauthoryear{Grimmond}{Grimmond}{2007}]%
        {grimmond2007urbanization}
\bibfield{author}{\bibinfo{person}{Sue~UE Grimmond}.}
  \bibinfo{year}{2007}\natexlab{}.
\newblock \showarticletitle{Urbanization and global environmental change: local
  effects of urban warming}.
\newblock \bibinfo{journal}{\emph{Geographical Journal}} \bibinfo{volume}{173},
  \bibinfo{number}{1} (\bibinfo{year}{2007}), \bibinfo{pages}{83--88}.
\newblock


\bibitem[\protect\citeauthoryear{Hansen, Bi, Saniotis, and Nitschke}{Hansen
  et~al\mbox{.}}{2013}]%
        {hansen2013vulnerability}
\bibfield{author}{\bibinfo{person}{Alana Hansen}, \bibinfo{person}{Linda Bi},
  \bibinfo{person}{Arthur Saniotis}, {and} \bibinfo{person}{Monika Nitschke}.}
  \bibinfo{year}{2013}\natexlab{}.
\newblock \showarticletitle{Vulnerability to extreme heat and climate change:
  is ethnicity a factor?}
\newblock \bibinfo{journal}{\emph{Global health action}} \bibinfo{volume}{6},
  \bibinfo{number}{1} (\bibinfo{year}{2013}), \bibinfo{pages}{21364}.
\newblock


\bibitem[\protect\citeauthoryear{Harlan, Brazel, Prashad, Stefanov, and
  Larsen}{Harlan et~al\mbox{.}}{2006}]%
        {harlan2006neighborhood}
\bibfield{author}{\bibinfo{person}{Sharon~L Harlan}, \bibinfo{person}{Anthony~J
  Brazel}, \bibinfo{person}{Lela Prashad}, \bibinfo{person}{William~L
  Stefanov}, {and} \bibinfo{person}{Larissa Larsen}.}
  \bibinfo{year}{2006}\natexlab{}.
\newblock \showarticletitle{Neighborhood microclimates and vulnerability to
  heat stress}.
\newblock \bibinfo{journal}{\emph{Social science \& medicine}}
  \bibinfo{volume}{63}, \bibinfo{number}{11} (\bibinfo{year}{2006}),
  \bibinfo{pages}{2847--2863}.
\newblock


\bibitem[\protect\citeauthoryear{Harlan, Declet-Barreto, Stefanov, and
  Petitti}{Harlan et~al\mbox{.}}{2013}]%
        {harlan2013neighborhood}
\bibfield{author}{\bibinfo{person}{Sharon~L Harlan}, \bibinfo{person}{Juan~H
  Declet-Barreto}, \bibinfo{person}{William~L Stefanov}, {and}
  \bibinfo{person}{Diana~B Petitti}.} \bibinfo{year}{2013}\natexlab{}.
\newblock \showarticletitle{Neighborhood effects on heat deaths: social and
  environmental predictors of vulnerability in Maricopa County, Arizona}.
\newblock \bibinfo{journal}{\emph{Environmental health perspectives}}
  \bibinfo{volume}{121}, \bibinfo{number}{2} (\bibinfo{year}{2013}),
  \bibinfo{pages}{197--204}.
\newblock


\bibitem[\protect\citeauthoryear{Huang, Zhou, and Cadenasso}{Huang
  et~al\mbox{.}}{2011}]%
        {huang2011everyone}
\bibfield{author}{\bibinfo{person}{Ganlin Huang}, \bibinfo{person}{Weiqi Zhou},
  {and} \bibinfo{person}{ML Cadenasso}.} \bibinfo{year}{2011}\natexlab{}.
\newblock \showarticletitle{Is everyone hot in the city? Spatial pattern of
  land surface temperatures, land cover and neighborhood socioeconomic
  characteristics in Baltimore, MD}.
\newblock \bibinfo{journal}{\emph{Journal of environmental management}}
  \bibinfo{volume}{92}, \bibinfo{number}{7} (\bibinfo{year}{2011}),
  \bibinfo{pages}{1753--1759}.
\newblock


\bibitem[\protect\citeauthoryear{Huang, Li, Guo, Zheng, and Qi}{Huang
  et~al\mbox{.}}{2020}]%
        {huang2020assessing}
\bibfield{author}{\bibinfo{person}{Xiaojun Huang}, \bibinfo{person}{Yanyu Li},
  \bibinfo{person}{Yuhui Guo}, \bibinfo{person}{Dianyuan Zheng}, {and}
  \bibinfo{person}{Mingyue Qi}.} \bibinfo{year}{2020}\natexlab{}.
\newblock \showarticletitle{Assessing Urban Risk to Extreme Heat in China}.
\newblock \bibinfo{journal}{\emph{Sustainability}} \bibinfo{volume}{12},
  \bibinfo{number}{7} (\bibinfo{year}{2020}), \bibinfo{pages}{2750}.
\newblock


\bibitem[\protect\citeauthoryear{Kianmehr and Lim}{Kianmehr and Lim}{2022}]%
        {kianmehr2022quantifying}
\bibfield{author}{\bibinfo{person}{Ayda Kianmehr} {and}
  \bibinfo{person}{Theodore~C Lim}.} \bibinfo{year}{2022}\natexlab{}.
\newblock \showarticletitle{Quantifying interactive cooling effects of
  morphological parameters and vegetation-related landscape features during an
  extreme heat event}.
\newblock \bibinfo{journal}{\emph{Climate}} \bibinfo{volume}{10},
  \bibinfo{number}{4} (\bibinfo{year}{2022}), \bibinfo{pages}{60}.
\newblock


\bibitem[\protect\citeauthoryear{Kianmehr, Lim, and Li}{Kianmehr
  et~al\mbox{.}}{2023}]%
        {kianmehr2023comparison}
\bibfield{author}{\bibinfo{person}{Ayda Kianmehr}, \bibinfo{person}{Theodore~C
  Lim}, {and} \bibinfo{person}{Xiaojiang Li}.} \bibinfo{year}{2023}\natexlab{}.
\newblock \showarticletitle{Comparison of different spatial temperature data
  sources and resolutions for use in understanding intra-urban heat variation}.
\newblock \bibinfo{journal}{\emph{Sustainable Cities and Society}}
  \bibinfo{volume}{96} (\bibinfo{year}{2023}), \bibinfo{pages}{104619}.
\newblock


\bibitem[\protect\citeauthoryear{Mallen, Stone, and Lanza}{Mallen
  et~al\mbox{.}}{2019}]%
        {mallen2019methodological}
\bibfield{author}{\bibinfo{person}{Evan Mallen}, \bibinfo{person}{Brian Stone},
  {and} \bibinfo{person}{Kevin Lanza}.} \bibinfo{year}{2019}\natexlab{}.
\newblock \showarticletitle{A methodological assessment of extreme heat
  mortality modeling and heat vulnerability mapping in Dallas, Texas}.
\newblock \bibinfo{journal}{\emph{Urban Climate}}  \bibinfo{volume}{30}
  (\bibinfo{year}{2019}), \bibinfo{pages}{100528}.
\newblock


\bibitem[\protect\citeauthoryear{Nielsen and Chuang}{Nielsen and
  Chuang}{2002}]%
        {nielsen2002quantum}
\bibfield{author}{\bibinfo{person}{Michael~A Nielsen} {and}
  \bibinfo{person}{Isaac Chuang}.} \bibinfo{year}{2002}\natexlab{}.
\newblock \bibinfo{title}{Quantum computation and quantum information}.
\newblock
\newblock


\bibitem[\protect\citeauthoryear{O’Neill, Zanobetti, and Schwartz}{O’Neill
  et~al\mbox{.}}{2005}]%
        {o2005disparities}
\bibfield{author}{\bibinfo{person}{Marie~S O’Neill},
  \bibinfo{person}{Antonella Zanobetti}, {and} \bibinfo{person}{Joel
  Schwartz}.} \bibinfo{year}{2005}\natexlab{}.
\newblock \showarticletitle{Disparities by race in heat-related mortality in
  four US cities: the role of air conditioning prevalence}.
\newblock \bibinfo{journal}{\emph{Journal of urban health}}
  \bibinfo{volume}{82}, \bibinfo{number}{2} (\bibinfo{year}{2005}),
  \bibinfo{pages}{191--197}.
\newblock


\bibitem[\protect\citeauthoryear{Pachauri, Allen, Barros, Broome, Cramer,
  Christ, Church, Clarke, Dahe, Dasgupta, et~al\mbox{.}}{Pachauri
  et~al\mbox{.}}{2014}]%
        {pachauri2014climate}
\bibfield{author}{\bibinfo{person}{Rajendra~K Pachauri},
  \bibinfo{person}{Myles~R Allen}, \bibinfo{person}{Vicente~R Barros},
  \bibinfo{person}{John Broome}, \bibinfo{person}{Wolfgang Cramer},
  \bibinfo{person}{Renate Christ}, \bibinfo{person}{John~A Church},
  \bibinfo{person}{Leon Clarke}, \bibinfo{person}{Qin Dahe},
  \bibinfo{person}{Purnamita Dasgupta}, {et~al\mbox{.}}}
  \bibinfo{year}{2014}\natexlab{}.
\newblock \bibinfo{booktitle}{\emph{Climate change 2014: synthesis report.
  Contribution of Working Groups I, II and III to the fifth assessment report
  of the Intergovernmental Panel on Climate Change}}.
\newblock \bibinfo{publisher}{Ipcc}.
\newblock


\bibitem[\protect\citeauthoryear{Richard, Kosatsky, and Renouf}{Richard
  et~al\mbox{.}}{2011}]%
        {richard2011correlates}
\bibfield{author}{\bibinfo{person}{Lucie Richard}, \bibinfo{person}{Tom
  Kosatsky}, {and} \bibinfo{person}{Annie Renouf}.}
  \bibinfo{year}{2011}\natexlab{}.
\newblock \showarticletitle{Correlates of hot day air-conditioning use among
  middle-aged and older adults with chronic heart and lung diseases: the role
  of health beliefs and cues to action}.
\newblock \bibinfo{journal}{\emph{Health education research}}
  \bibinfo{volume}{26}, \bibinfo{number}{1} (\bibinfo{year}{2011}),
  \bibinfo{pages}{77--88}.
\newblock


\bibitem[\protect\citeauthoryear{Theoharatos, Pantavou, Mavrakis, Spanou,
  Katavoutas, Efstathiou, Mpekas, and Asimakopoulos}{Theoharatos
  et~al\mbox{.}}{2010}]%
        {theoharatos2010heat}
\bibfield{author}{\bibinfo{person}{George Theoharatos},
  \bibinfo{person}{Katerina Pantavou}, \bibinfo{person}{Anastasios Mavrakis},
  \bibinfo{person}{Anastasia Spanou}, \bibinfo{person}{George Katavoutas},
  \bibinfo{person}{Panos Efstathiou}, \bibinfo{person}{Periklis Mpekas}, {and}
  \bibinfo{person}{Dimosthenis Asimakopoulos}.}
  \bibinfo{year}{2010}\natexlab{}.
\newblock \showarticletitle{Heat waves observed in 2007 in Athens, Greece:
  synoptic conditions, bioclimatological assessment, air quality levels and
  health effects}.
\newblock \bibinfo{journal}{\emph{Environmental Research}}
  \bibinfo{volume}{110}, \bibinfo{number}{2} (\bibinfo{year}{2010}),
  \bibinfo{pages}{152--161}.
\newblock


\bibitem[\protect\citeauthoryear{Uejio, Wilhelmi, Golden, Mills, Gulino, and
  Samenow}{Uejio et~al\mbox{.}}{2011}]%
        {uejio2011intra}
\bibfield{author}{\bibinfo{person}{Christopher~K Uejio},
  \bibinfo{person}{Olga~V Wilhelmi}, \bibinfo{person}{Jay~S Golden},
  \bibinfo{person}{David~M Mills}, \bibinfo{person}{Sam~P Gulino}, {and}
  \bibinfo{person}{Jason~P Samenow}.} \bibinfo{year}{2011}\natexlab{}.
\newblock \showarticletitle{Intra-urban societal vulnerability to extreme heat:
  the role of heat exposure and the built environment, socioeconomics, and
  neighborhood stability}.
\newblock \bibinfo{journal}{\emph{Health \& place}} \bibinfo{volume}{17},
  \bibinfo{number}{2} (\bibinfo{year}{2011}), \bibinfo{pages}{498--507}.
\newblock


\bibitem[\protect\citeauthoryear{Xu, Sheffield, Su, Wang, Bi, and Tong}{Xu
  et~al\mbox{.}}{2014}]%
        {xu2014impact}
\bibfield{author}{\bibinfo{person}{Zhiwei Xu}, \bibinfo{person}{Perry~E
  Sheffield}, \bibinfo{person}{Hong Su}, \bibinfo{person}{Xiaoyu Wang},
  \bibinfo{person}{Yan Bi}, {and} \bibinfo{person}{Shilu Tong}.}
  \bibinfo{year}{2014}\natexlab{}.
\newblock \showarticletitle{The impact of heat waves on children’s health: a
  systematic review}.
\newblock \bibinfo{journal}{\emph{International journal of biometeorology}}
  \bibinfo{volume}{58}, \bibinfo{number}{2} (\bibinfo{year}{2014}),
  \bibinfo{pages}{239--247}.
\newblock


\bibitem[\protect\citeauthoryear{Yu, Mengersen, Wang, Ye, Guo, Pan, and
  Tong}{Yu et~al\mbox{.}}{2012}]%
        {yu2012daily}
\bibfield{author}{\bibinfo{person}{Weiwei Yu}, \bibinfo{person}{Kerrie
  Mengersen}, \bibinfo{person}{Xiaoyu Wang}, \bibinfo{person}{Xiaofang Ye},
  \bibinfo{person}{Yuming Guo}, \bibinfo{person}{Xiaochuan Pan}, {and}
  \bibinfo{person}{Shilu Tong}.} \bibinfo{year}{2012}\natexlab{}.
\newblock \showarticletitle{Daily average temperature and mortality among the
  elderly: a meta-analysis and systematic review of epidemiological evidence}.
\newblock \bibinfo{journal}{\emph{International journal of biometeorology}}
  \bibinfo{volume}{56}, \bibinfo{number}{4} (\bibinfo{year}{2012}),
  \bibinfo{pages}{569--581}.
\newblock


\end{thebibliography}

\end{document}